\documentclass{article}

\usepackage{arxiv}

\usepackage[utf8]{inputenc} % allow utf-8 input
\usepackage[T1]{fontenc}    % use 8-bit T1 fonts
\usepackage{hyperref}       % hyperlinks
\usepackage{url}            % simple URL typesetting
\usepackage{booktabs}       % professional-quality tables
\usepackage{amsfonts}       % blackboard math symbols
\usepackage{nicefrac}       % compact symbols for 1/2, etc.
\usepackage{microtype}      % microtypography
\usepackage{lipsum}		% Can be removed after putting your text content
\usepackage{graphicx}
\usepackage{natbib}
\usepackage{doi}
\usepackage{caption}
\usepackage{subcaption}
\usepackage{booktabs,chemformula}
\usepackage{multirow}

\title{Cascaded Deep Hybrid Models for Multistep Household Energy Consumption Forecasting}

\author{$^a$Lyes  SAAD SAOUD, ~$^a$Hasan AlMarzouqi, ~$^b$Ramy  Hussein  \\
	$^a$Electrical Engineering and Computer Science Department, Khalifa University, PO Box 127788, Abu Dhabi, UAE.\\
	$^b$Radiological Sciences Laboratory, Stanford University, 1201 Welch Road, Stanford, California 94305-5488, USA \\
        \texttt{lyes.saoud@ku.ac.ae, hasan.almarzouqi@ku.ac.ae, ramyh@stanford.edu}
}
\renewcommand{\undertitle}{Under consideration at Pattern Recognition Letters}
\renewcommand{\shorttitle}{Cascaded Deep Hybrid Models for Multistep Household Energy Consumption Forecasting}

\hypersetup{
pdftitle={Cascaded Deep Hybrid Models for Multistep Household Energy Consumption Forecasting},
pdfsubject={q-bio.NC, q-bio.QM},
pdfauthor={Lyes  SAAD SAOUD, Hasan AlMarzouqi, and Ramy  Hussein},
pdfkeywords={Power Consumption Forecasting, Conventional Neural Networks, Long Short Term Memory, Stationary Wavelet Transform},
}

\begin{document}
\maketitle

\begin{abstract}
Sustainability requires increased energy efficiency with minimal waste. The future power systems should thus provide high levels of flexibility in controlling energy consumption. Precise projections of future energy demand/load at the aggregate and on the individual site levels are of great importance for decision-makers and professionals in the energy industry. Forecasting energy loads has become more advantageous for energy providers and customers, allowing them to establish an efficient production strategy to satisfy demand. This paper introduces two hybrid cascaded models for forecasting multistep household power consumption in different resolutions. The first model integrates Stationary Wavelet Transform (SWT), as an efficient signal preprocessing technique, with Convolutional Neural Networks and Long Short Term Memory (LSTM) units. The second hybrid model combines SWT with a self-attention based neural network architecture named transformer. The major constraint of using time-frequency analysis methods such as SWT in multistep energy forecasting problems is that they require sequential signals, making signal reconstruction problematic in multistep forecasting applications. The cascaded models can efficiently address this problem by using recursive outputs. Experimental results show that the proposed hybrid models achieve superior prediction performance compared to existing multi-step power consumption prediction methods. The results will pave the way for more accurate and reliable forecasting of household power consumption.
\end{abstract}

% \end{frontmatter}

%\linenumbers

%% main text
\section{Introduction}
Recent rapid economic growth and development have increased the global use of electricity worldwide \citep{9141253}. According to the International Energy Agency (IEA) World Energy Outlook 2019 report, the global energy demand rises by 2.1\% per year, and the trend will continue until 2040, which is twice as much as expected in the started policies scenario \citep{IEA}. The housing industry accounts for 27\% of the world's demand for power, which considerably impacts the overall consumption of electricity \citep{NEJAT2015843}. Effective forecasting of energy consumption could potentially maintain higher stability of power supplies, especially in hybrid power generation systems \citep{KIM201972}. In recent decades, many models have been developed to predict energy consumption in several building types \citep{en10101525}.
\begin{figure*}%[ht]
\centering
    \includegraphics[width=0.95\linewidth]{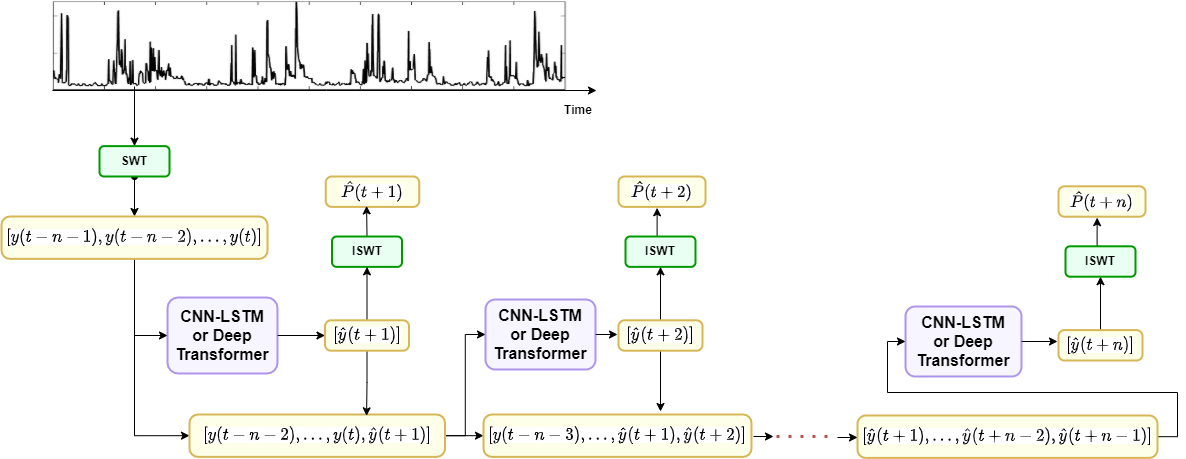}\
\caption{ The cascaded CNN-LSTM/Transformer -SWT hybrid model for Multistep household power consumption forecasting}
\label{fig: a1}
\end{figure*}
Forecasting energy consumption is essential for the accurate assessment of the economic, social, and environmental factors that cause recurrent fluctuations in energy demands. Smart grids equipped with the latest machine learning technologies will offer more flexibility in the generation and distribution of energy. For instance, smart grids can dynamically react to the energy demand changes  and effectively distribute energy generated from renewable sources \citep{SIANO2014461}. Power usage prediction has recently been the focus of researchers and manufacturers as this is a key and important task for economically effective operation and controls \citep{10.1007/978-981-13-9714-1_20}. Accurate load forecasting ensures safe and reliable power systems operation \citep{ROLDANBLAY201338}. With the complexity of the power systems, predicting the energy consumption of modern buildings is deemed challenging. The prediction of household energy usage is often affected by several elements, such as the type of electrical equipment and appliances inside the house, their geographical locations, and their activity time \citep{AHMAD2014102}.

Based on the time horizon, electrical load forecasting models may be classified into four prime categories: very short-term, short-term, medium-term, and long-term prediction \citep{ESKANDARI2021107173}. Numerous load forecasting methods have been presented  to address such forecasting problems. These methods are categorized into three main types: (i) engineering approaches, which compute thermal dynamics and energy behavior for the entire buildings, (ii) statistical techniques, which explore the relationship between energy consumption and other elements such as climatic data and occupation, and (iii) machine learning techniques, which focus more on learning the distinguishable energy usage patterns from the historical data. However, both engineering and statistical methods were found to be unreliable and less generalizable in real-life conditions, especially when tested on unseen data at a new site \citep{ZHAO20123586, KABOLI2016857}. 

Statistical techniques were used mainly for predicting energy demands. For instance, the autoregressive integrated moving average (ARIMA) model was applied to forecast power consumption (\cite{ en4081246}). \cite{ en14185831} have evaluated three different models, seasonal ARIMA (SARIMA), XGBOOST, and LSTM, for forecasting power consumption. Their results showed that the XGBOOST model is able to achieve higher prediction performance in the one-minute ahead forecasting problem compared to other methods. In an attempt to solve the volatility problem in energy consumption data, singular spectrum analysis (SSA) and parallel LSTM (pLSTM) neural networks were integrated to better forecast power consumption \citep{ JIN2022101442}.

In the past few years, several machine learning-based energy consumption prediction models have been proposed. For instance, a model based on LSTM and a sine cosine optimization algorithm was proposed by \citep{SOMU2020114131}, which improved the short-term forecasting accuracy by around 12\% and long-term forecasting by 60\%. A dilated convolutional neural network (DCNN) was also combined with bidirectional LSTM (BiLSTM) to efficiently control power usage in integrated local energy systems between consumers and suppliers (\cite{ KHAN2021107023}). \cite{HU2020598} combined echo state network, bagging, and a differential evolution algorithm to improve energy consumption forecasting accuracy. A combination of progress learning and deep learning was introduced to address nonlinear and complex energy consumption forecasting. Their proposed model improved the root mean squared error (RMSE) by more than 17\% compared to traditional backpropagation neural networks \citep{LIU2020109675}. An ensemble hybrid model was also proposed to predict the cyclic and stochastic components of building energy in \citep{ZHANG2020117531}. Obtained results showed that the cyclic features can improve the prediction RMSE by 30\%. A hybrid forecasting model based on data mining techniques and the Box-Jenkins method was presented to capture the nonlinear and complex patterns in peak load and energy demand data \citep{KAZEMZADEH2020117948}. A hybrid model based on empirical mode decomposition and extreme gradient boosting was proposed to deal with the high nonlinearity of energy consumption prediction \citep{LU2020117756}. Convolutional networks were integrated with Gated Recurrent Units (GRU) in \citep{9141253}, and with LSTM \citep{KIM201972} to better leverage the spatio-temporal features and improve the overall prediction performance. Experiments showed that hybrid CNN-GRU networks are 50\% more accurate than hybrid CNN-LSTM. To reduce volatility and expand data dimensionality, the stationary wavelet transform (SWT) was combined with LSTM units to forecast energy consumption \citep{8880605}. Results revealed that using SWT for data pre-processing can remarkably improve LSTM forecasting performance. Transformer networks were recently proposed to overcome LSTM's parallelization problem \citep{NIPS2017_3f5ee243}. Transformer-based models can efficiently capture complex time-series data dynamics that are difficult to extract with traditional sequence models like recurrent neural networks (RNNs) \citep{https://doi.org/10.48550/arxiv.2001.08317}.

Energy usage patterns in individual households are typically irregular due to a variety of factors, such as weather and holidays. As a result, methods that anticipate energy usage solely using energy consumption statistics are not accurate enough. Therefore, including other observations (whether the observed point is an anomaly, a change point, or a pattern) may help boost the prediction performance \citep{NEURIPS2019_6775a063}. The similarities between distinct data encoding variables in transformers (e.g., queries and keys) are determined based on their point-specific values rather than taking local contexts into account \citep{10.1007/978-3-030-63836-8_51}. This problem might be solved by either introducing an efficient attention mechanism to replace the traditional self-attention module (SAM) of transformers, such as the Spring Time Warping Matrix \citep{10.1007/978-3-030-63836-8_51}, or externally providing additional information about the surroundings of the observed point to the transformer \citep{8880605}. The proposed forecasting strategy in this paper is based on the latter approach.
The SWT is effective when combined with other machine learning models, and it has been proven to be an efficient pre-processing approach that decomposes a given time-series signal into high- and low-frequency sub-signals, revealing distinct patterns from the actual observations \citep{8880605}. In a previous paper \citep{9672113}, a hybrid model based on the transformers and SWT was proposed to forecast one step ahead energy consumption. It has been shown that integrating SWT with transformer networks is 48\% more accurate than the SWT’s hybridization with LSTM. A fundamental drawback of using SWTs in multistep forecasting is that they require sequential signals for reconstruction, but  sequences of future samples don't exist when solving multistep prediction problems. 

Despite great efforts to improve household energy consumption forecasting, previous methods still have some limitations that need to be addressed. (i) In multistep forecasting problems that are based on decomposition/reconstruction, it is not possible to reconstruct the forecasted signal from its wavelet coefficients due to the unavailability of future data, i.e., SWT coefficients need to be forecasted first in order to reconstruct the signal. (ii) Deep learning models require a large amount of training data and high computational power. 

To fill the aforementioned research gap, we propose hybrid cascaded deep learning-based models that combine either the CNN-LSTM or Transformer with the SWT to accurately predict household energy consumption. This is an extension of previous studies that showed great potential for integrating SWT with deep learning \citep{8880605, 9672113}. The proposed models extend the usage of the SWT-based hybrid predictors to multistep forecasting problems, in which future data are needed for signal reconstruction.
    
The main contributions of this work can be summarized as follows:
\begin{itemize}
  \item We developed two hybrid SWT-Deep learning models to forecast multistep residential energy consumption. The deep learning models predict the SWT's energy 
consumption-related features and feed them recursively along the forecasting horizon to predict the next steps. This combination helps in describing the irregular patterns in univariate household energy data and resolves the multistep SWT reconstruction issue. This is the first time, to the best of our knowledge, that a combination of SWT and deep learning models is used to create a multistep energy consumption prediction model.
  \item Experiments based on a benchmark energy consumption dataset and for different time resolutions reveal that the proposed SWT-deep learning models can effectively forecast home energy usage and outperform concurrent and previous prediction methods.
\end{itemize}

\begin{figure*}[]
\centering
    \includegraphics[scale=1, width=0.950\linewidth]{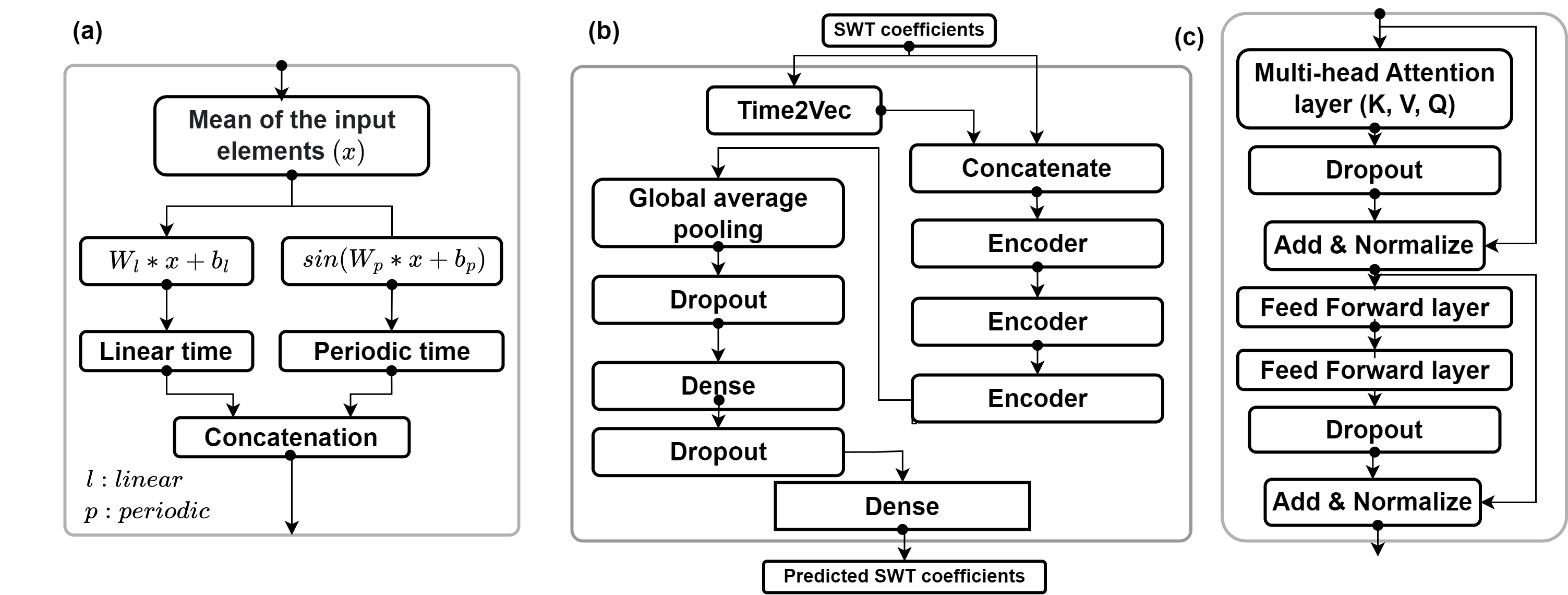}\
\caption{(a) Time2Vec block, (b) the proposed deep transformer-based SWT forecasting model, and (c) the encoder block}
\label{fig: a3}
\end{figure*}
\begin{figure}
\centering
\includegraphics[scale=0.25]{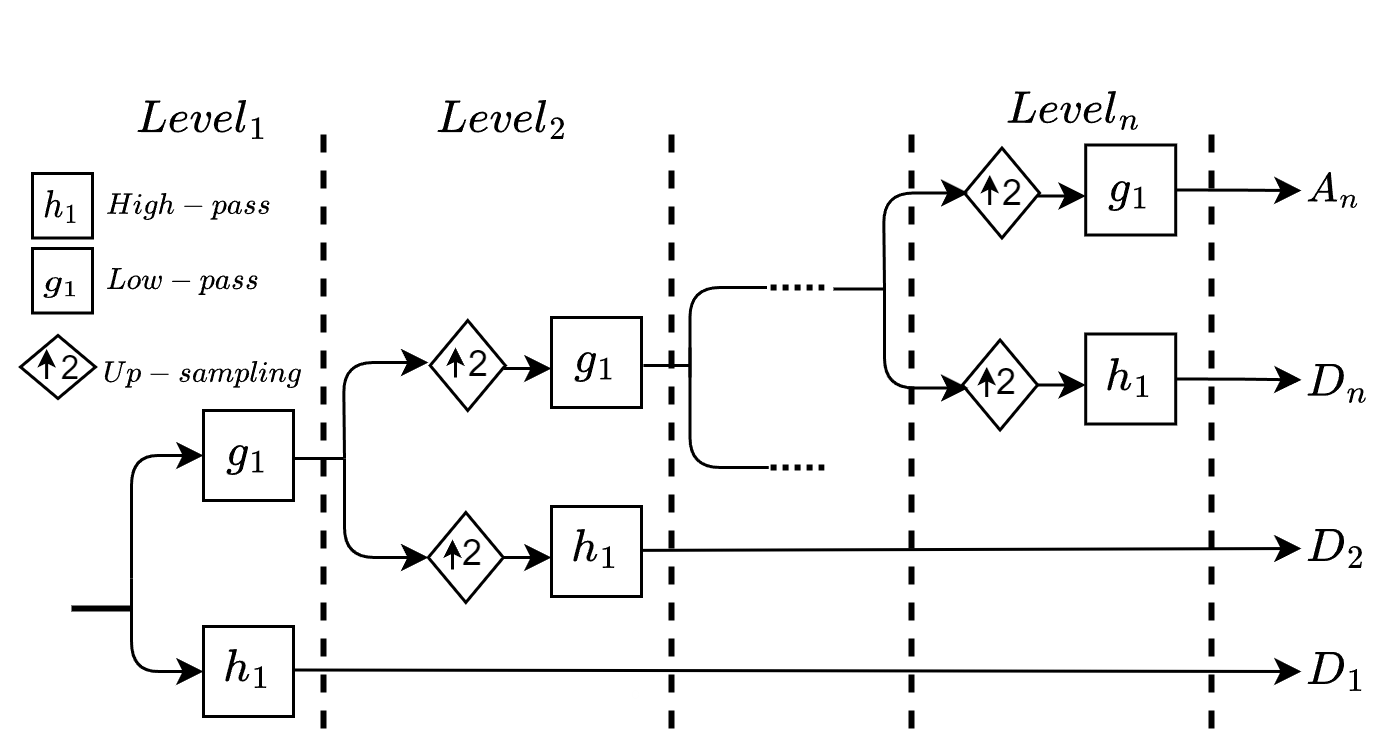}
\captionof{figure}{N-level SWT decomposition}
\label{fig: swt}
\end{figure}

\section{The Proposed Model}
We propose two hybrid prediction methods based on the stationary wavelet transform and two deep neural network architectures (CNN-LSTM and Transformer) for forecasting household energy consumption. First, the univariate energy input data is decomposed into approximation and detail subbands using SWT to extract the distinguishable patterns needed for accurate predictions. Second, the wavelet sub-bands are supplied to either a CNN-LSTM network or a deep transformer network in order to forecast the next wavelet sub-bands.   Finally, the inverse SWT is applied to the outputs of deep learning models to reconstruct the predicted household energy consumption. A schematic diagram of the proposed prediction methods is shown in Figure ~\ref{fig: a1}.

\subsection{The stationary wavelet transform } 
The wavelet transform divides the energy consumption signal into levels; the resultant subsignals are half the length of the approximation signal at the previous wavelet level. The stationary wavelet transform (SWT) \citep{Nason1995}, shown in Figure~\ref{fig: swt}, was introduced to address shift-invariance and non-redundancy limitations of discrete wavelet transform (DWT) \citep{article, GUVEN2020106678}. Instead of down-sampling the data after applying the low-pass or high-pass filters in the DWT, SWT adjusts the filters at each decomposition level by padding zeroes \citep{HADIYAN2020100401}. The SWT eliminates the down-sampling operator from the standard DWT implementation. The generated subsignals in the SWT have the same length as the source signal, which is a desirable attribute for the proposed forecasting model. Using the SWT offers three advantages: 1) Unlike conventional  wavelets, the SWT sub-signals have the same length as the original signal, which makes it most suitable for neural networks, 2) the SWT sub-bands include crucial information needed for accurate prediction and, 3) the SWT has a low computational cost, which makes it a perfect candidate for real-time analyses and predictions \citep{ZHOU2020125127}. Therefore, we use the SWT to produce identifiable low- and high-frequency components (known as approximations and details) that well represent the energy consumption time-series data. These components are then used as inputs to either the CNN-LSTM architecture or the deep transformer for energy consumption prediction.
The SWT approximation sub-band depicts the time-series overall trend, and the detail sub-band shows small series deviations. The SWT deconstructs the time series using a hierarchical mix of low-pass and high-pass wavelet filters, allowing high and low frequencies to be separated. The breakdown can be shown as a dyadic tree form \citep{TALAAT2020117087} (see Figure ~\ref{fig: swt}).  
We examined four wavelet families and five decomposition levels, and our results reveal that the Daubechies (db1) with three levels produce minimum prediction errors. It should be noted that the inverse SWT (ISWT) only needs the details and the last approximation signals to reconstruct the original signal, but we have noticed that using all subsignals improves the results. Therefore, we have used all six wavelet features to train the networks. The total number of wavelet features used is $6 \times N$, where $N$ is the total number of samples.  

\subsection{SWT-deep learning model} 
The proposed hybrid deep learning-based energy consumption methods integrate SWT with either CNN-LSTM or Transformer network.

First, we decomposed the energy consumption dataset into sublevels using SWT. Then, we applied the deep learning models to forecast the next SWT coefficients $\hat{y}(t+1)$ using the $n$ past values of the SWT coefficients $[y(t-n), y(t-n+1), \ldots, y(t)]$. The ISWT was applied to the predicted coefficients to have the one step ahead  predicted energy value $\hat{P}(t+1)$. The predicted SWT coefficients $\hat{y}(t+1)$ were also supplied together with the shifted SWT coefficients vector to the deep learning models to forecast the next SWT coefficients $\hat{y}(t+2)$. This procedure is recursively applied until the prediction of all values of the entire horizon is achieved. Note that we trained the network to forecast one step ahead, and then it was fine-tuned sequentially for the next steps. 

We examine the performance of both CNN-LSTM and Transformer networks for automated feature learning and prediction. LSTM is a type of recurrent neural network that may learn long-term relationships, particularly in sequence prediction issues. A cell, an input gate, an output gate, and a forget gate comprise a typical LSTM unit. The following are the  equations describing a forward pass of the LSTM cell \citep{199317, 10.1162/neco.1997.9.8.1735}:
\begin{equation}
  i_{t} =\sigma _{1}(W_{i}x_{t}+U_{i}h_{t-1}+b_{i})
\end{equation}
\begin{equation}
  f_{t} =\sigma _{1}(W_{f}x_{t}+U_{f}h_{t-1}+b_{f})
\end{equation}
\begin{equation}
  o_{t} =\sigma _{1}(W_{o}x_{t}+U_{o}h_{t-1}+b_{o})
\end{equation}
\begin{equation}
  {\tilde {c}}_{t} =\sigma _{2}(W_{c}x_{t}+U_{c}h_{t-1}+b_{c})
\end{equation}
\begin{equation}
  c_{t} =f_{t}\odot c_{t-1}+i_{t}\odot {\tilde {c}}_{t}
\end{equation}
\begin{equation}
  h_{t} =o_{t}\odot \sigma _{2}(c_{t})
\end{equation}
where $x_t$ and $h_t$ are the cell's hidden input and output at time step $t$, respectively. The matrices $W_q$ and $U_q$ hold the weights of the input and recurrent connections, respectively. $b_q,~q=\{i, f, o, c\}$ is the bias. The hidden output $h_t$ is computed using equations (1) to (6), where  $i$ is the input gate, $o$ is the output gate, $f$ is the forget gate, and $c$ is the  memory cell. $\sigma _{1}$ and $\sigma _{2}$ are the sigmoid and  hyperbolic tangent activation functions, respectively. The initial values are $c_{0}=0$ and $h_{0}=0$ and the operator $\odot$  denotes the Hadamard product. 

The proposed CNN-LSTM-SWT network uses three 1D convolutional layers with Rectified Linear Unit (ReLU) activation function. The CNN layers are followed by two layers of LSTM cells and two fully connected layers. The fully connected layer uses ReLU and linear activation functions. We also used dropouts in the LSTM and fully connected layers to address model overfitting.

Figure ~\ref{fig: a3} depicts the schematic diagram of the transformer-based SWT decomposition and reconstruction model for multistep time series prediction. The model contains one Time2Vec block followed by three encoder blocks and one global average pooling, then two sequential dropout-dense layers. The transformer is a deep learning architecture that relies heavily on attention mechanisms to analyze sequential input data. Unlike previous sequence models, the transformer relaxes the need for using convolutional or recurrent neural networks, it employs stacked multi-head self-attention units and fully connected layers. Each layer begins with a multi-head self-attention layer, followed by two feedforward levels. Both multi-head attention and feedforward layers are followed by dropout and Add\&Normlize layers. The proposed prediction method takes advantage of the transformer’s encoder structure only while ditching the decoders completely. The SWT coefficients are passed through the Time2Vec block before going to the encoders (Figure ~\ref{fig: a3}). The Time2Vec \citep{ https://doi.org/10.48550/arxiv.1907.05321, 9194535} is an expanded and learnable version of the transformer's initial positional encoding layer. It permits learning the input frequencies rather than utilizing a fixed representation (see Figure ~\ref{fig: a1}(a)). The Time2Vec operation implements the following equation (\cite{ https://doi.org/10.48550/arxiv.1907.05321}):

\begin{equation}
    \label{eqn:equation1}
    Time2Vec\left( \tau  \right)\left[ i\right]=
    \begin{cases}
     {{\omega }_{i}}\tau +{{\varphi }_{i}}~~~~~~~~~~~~~~~\text{if}~~i=0\\
    \sin\left( {{\omega }_{i}}\tau +{{\varphi }_{i}} \right),~~~~\text{if}~~~1\le i\le k 
    \end{cases}
\end{equation}
where, $Time2Vec\left( \tau  \right)\left[ i \right]$  is the $i^{th}$ element of $Time2Vec\left( \tau  \right)$ that has k elements, and ${{\omega }_{i}}$ and ${{\varphi }_{i}}$ are learnable parameters.

\begin{figure}
\centering
\includegraphics[scale=0.2]{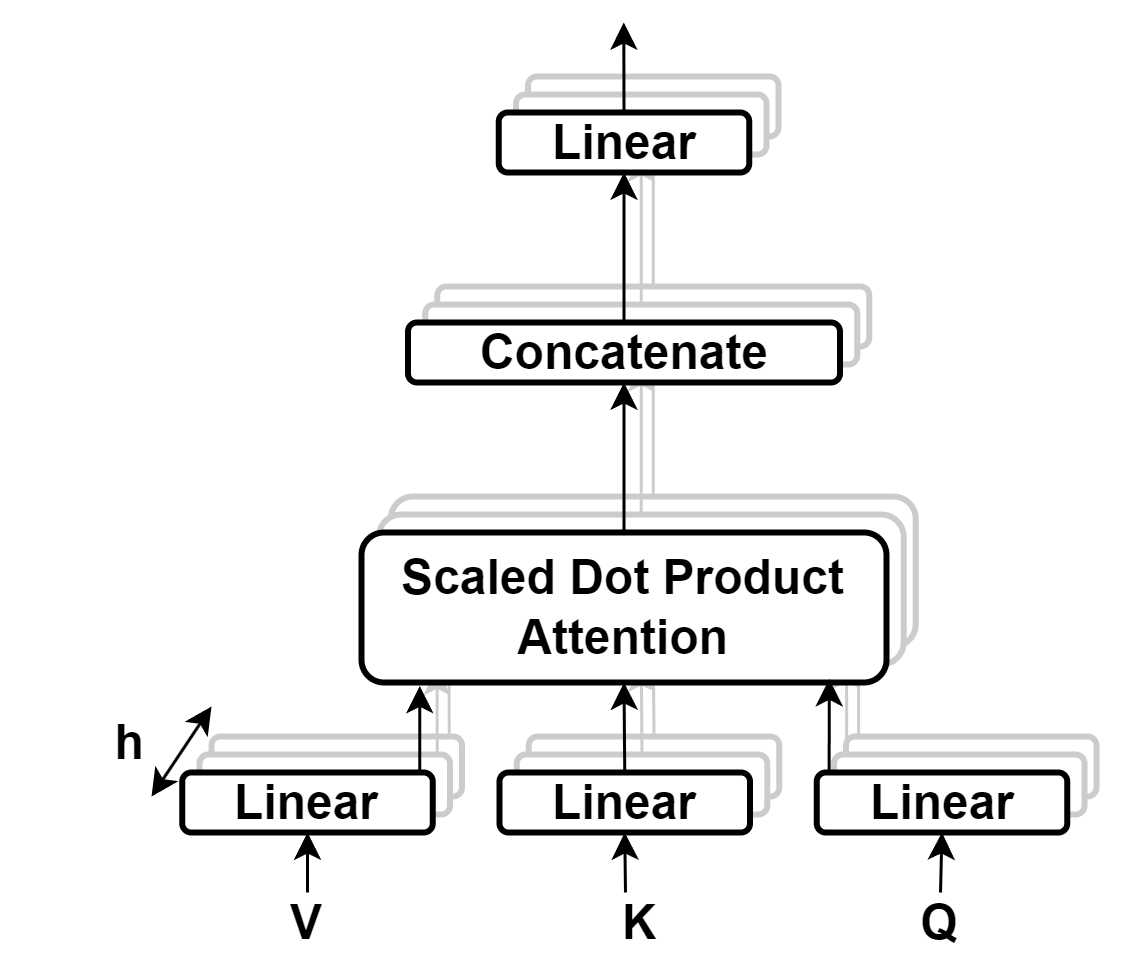}
\captionof{figure}{Multi-head attention. V, K, and Q are learned linear projections of input data, and h is the number of parallel attention layers }
\label{fig: MHA}
\end{figure}

The encoder comprises a set of sub-encoders that handle each layer's input sequentially. Each encoder layer generates encodings of essential information based on the portions of the inputs that are significant to one another (Figure 2(c)). Next, the encodings are passed into the next encoder layer, which uses attention mechanisms to compute the relevance of each input. Encoder layers include residual connections, layer normalization operations, and feedforward networks for further output processing. Each multi-head attention system has three learnable weights: query weights Q, key weights K, and value weights V \citep{LIU2020109675}. More specifically, the transformer's multi-head attention module conducts its computations in parallel (Figure~\ref{fig: MHA}). The attention module conducts an attention mechanism repeatedly in parallel. A single attention module output is given by \citep{LIU2020109675}:

\begin{equation}
Att_{j}=softmax( \frac{QK}{\sqrt{{{d}_{k}}}})V
\end{equation}
where: $j=1,\cdots ,h$,and $d_k$ is the dimension of query and key vectors.
The multi-head attention score is the concatenation of the output of $h$ heads given by Eq. ~\ref{eq: mha} multiplied with learnable projection parameters $W$, i.e.:

\begin{equation}
\label{eq: mha}
MultiheadAtt.\,=Concat\left( Att{{.}_{1}},\cdots ,Att{{.}_{h}} \right)W
\end{equation}
With multi-head attention, the transformer may encode multiple relationships and nuances for each input variable. The parameters used to set up our networks are given as follows:  batch size = 32, and the default Keras kernel initializer, i.e., Glorot uniform, was used to initialize the weights. Twelve parallel attention layers, with $Q=K=V=256$, are adopted in the proposed method. The total number of trainable parameters of the two proposed models are 340,294 parameters for the Transformer-SWT model and 422,358 parameters for the CNN-LSTM-SWT model.

\section{Results and discussions}
\textbf{Data description:}
The University of California Irvine (UCI) machine learning repository provides a power consumption dataset with 2,075,259 measurements sampled at one minute and collected between December 2006 and November 2010 at the house of Sceaux (Paris, France) \citep{misc_individual_household_electric_power_consumption_235}. Our forecasting paper focuses on global active power (i.e. we use global active power time series in the forecasting process). It should be noted that this dataset was used by many researchers to validate their models’ performance (e.g., \citep{ MOCANU201691, 7793413}). For a fair comparison with other existing studies, we have used the same strategy applied in (\citep{ MOCANU201691, 7793413}) to test the performance of our proposed systems. The first three years of the dataset were utilized to train the models, and the remaining dataset was used to test the models. The root mean squared error (RMSE), the mean absolute error (MAE), and the mean absolute percentage error (MAPE) are chosen as metrics for the models' evaluation.	

This paper proposes novel deep learning solutions to forecast multistep energy consumption for several time scales. The performance of the developed model was compared with the following models: Auto-Regressive Integrated Moving Average (ARIMA), Support Vector Regression (SVR), random forest, a multi-input multi-output real-valued neural network (MIMO-RVNN), a wavelet network, a Radial basis function (RBF) network,  a Long Short Term Memory (LSTM) model, and a hybrid convolutional neural network-LSTM model (CNN-LSTM). In addition, results were compared with recently published studies addressing the same problem \citep{9141253, KIM201972, MOCANU201691, 8945363, 7793413}.  The model settings used in our experiments are given as follows:

1) ARIMA model: the grid search algorithm \citep{brownlee2017introduction} was used to find the optimal parameters $(p, d, q) =(0, 1, 1)$.

2) SVR model: We used the common Gaussian kernel as a kernel for the SVR model \citep{doi:10.1080/07474938.2010.481556}. 

3) The random forest model: We used the same setting in \citep{en15134880}

4) The LSTM model: We used the same setting in \citep{Nazir_2021}

5) The MIMO-RVNN, wavelet network, and RBF network: These networks have one hidden layer. The optimal number of neurons in the hidden layer was found empirically and is equal to 100, 10, and 10 neurons for the MIMO-RVNN, wavelet network, and RBF network respectively.

\begin{table}[]
\centering
\footnotesize
\caption{Comparison results of the proposed model with other existing machine learning models }
%%\begin{tabular}{p{2.4cm}|p{1.3cm}p{0.9cm}p{0.9cm}c}
\begin{tabular}
{p{5cm}|p{1.8cm}p{1.8cm}p{1.8cm}p{1.8cm}}

\toprule
 Method                                               & Resolution & RMSE (kW)   & MAE (kW)     & MAPE (\%) \\
\midrule
\multirow{5}{*}{ARIMA}                         & Minutely   & 1.1542&0.6078&	22.39     \\
                                                      & Hourly     &1.0047&	0.8376&	155.0     \\
                                                      & Daily      & 0.3855 &	0.3067&	40.37     \\
                                                      & Weekly     &0.5498&	0.4658&	57.70     \\

\hline
\multirow{5}{*}{SVR}                    & Minutely   &0.7120 &0.4551 &69.73      \\
                                       & Hourly     &0.6942 &0.4927 &61.87       \\
                                    & Daily  &0.3571 &0.2709 &29.41  \\
                                         & Weekly  &0.1790 &0.1548 &16.17 
         \\
                                                      
\hline
\multirow{5}{*}{Random Forest}        & Minutely   &0.8242&	0.4909&	61.87     \\
                                       & Hourly    &0.8224 &0.6732 &113.7        \\
                                       & Daily       &0.3484 &0.2825 &38.19  \\
                                        & Weekly    &0.2062 &0.1494 &19.72        \\
\hline
\multirow{5}{*}{MIMO-RVNN}                         & Minutely   & 0.4879 & 0.3289 & 56.25     \\
                                                      & Hourly     & 0.9824 & 0.7229 & 105.1     \\
                                                      & Daily      & 0.5380 & 0.4316 & 57.92     \\
                                                      & Weekly     & 0.2466 & 0.1644 & 16.25     \\
\hline
\multirow{5}{*}{LSTM}                                 & Minutely   & 0.6150 & 0.4331 & 82.03     \\
                                                      & Hourly     & 0.6918 & 0.9268 & 107.4     \\
                                                      & Daily      & 0.3090 & 0.2306 & 26.26     \\
                                                      & Weekly     & 0.2965 & 0.1622 & 18.71     \\
\hline
\multirow{5}{*}{CNN-LSTM \citep{KIM201972}}           & Minutely   & 0.6114 & 0.3493 & 34.84     \\
                                                      & Hourly     & 0.5957 & 0.3317 & \textbf{32.83}     \\
                                                      & Daily      & 0.3221 & 0.2569 & 31.83     \\
                                                      & Weekly     & 0.3085 & 0.2382 & 31.84     \\
\hline
\multirow{5}{*}{Wavelet network}                      & Minutely   & 0.7479 & 0.5282 & 90.37     \\
                                                      & Hourly     & 0.7460 & 0.5563 & 86.81     \\
                                                      & Daily      & 0.7893 & 0.6183 & 83.00     \\
                                                      & Weekly     & 0.4697 & 0.5251 & 34.18     \\
\hline
\multirow{5}{*}{RBF network}                          & Minutely   & 0.7770 & 0.5301 & 85.85     \\
                                                      & Hourly     & 0.7519 & 0.5624 & 88.39     \\
                                                      & Daily      & 0.7565 & 0.6034 & 81.03     \\
                                                      & Weekly     & 0.2268 & 0.1673 & 16.93     \\
\hline
\multirow{5}{*}{Multistep CNN-LSTM-SWT (Proposed)}    & Minutely   & 0.4645 & 0.2060 & 33.99 \\
                                                      & Hourly     & 0.5102 & 0.3369 & 46.29     \\
                                                      & Daily      & 0.2793 & 0.2196 & 31.73     \\
                                                      & Weekly     & 0.1758 & 0.1261 & 12.43     \\
\hline
\multirow{5}{*}{Multistep Transformer-SWT (Proposed)} & Minutely   & \textbf{0.3929} & \textbf{0.1911} & \textbf{15.01}\\
                                                      & Hourly     & \textbf{0.4183} & \textbf{0.2637} & 35.04     \\
                                                      & Daily      & \textbf{0.2378} & \textbf{0.1832} & \textbf{21.90}     \\
                                                      & Weekly     & \textbf{0.1574} & \textbf{0.1106} & \textbf{11.72}   \\ 
\bottomrule
\end{tabular}
\label{table:2}
\end{table}

The proposed model architectures have been implemented in Keras\footnote{Chollet, F. et al. (2015). Keras. https://keras.io}  with Tensorflow backend \citep{199317} and the RMSProp learning algorithm, with a learning rate \(\eta=0.001\). As a preprocessing step, the SWT approximation and detail coefficients were normalized between 0 and 1 to speed up the learning and convergence of the proposed model. 

\begin{figure}[t]
\centering
\includegraphics[scale=0.3]{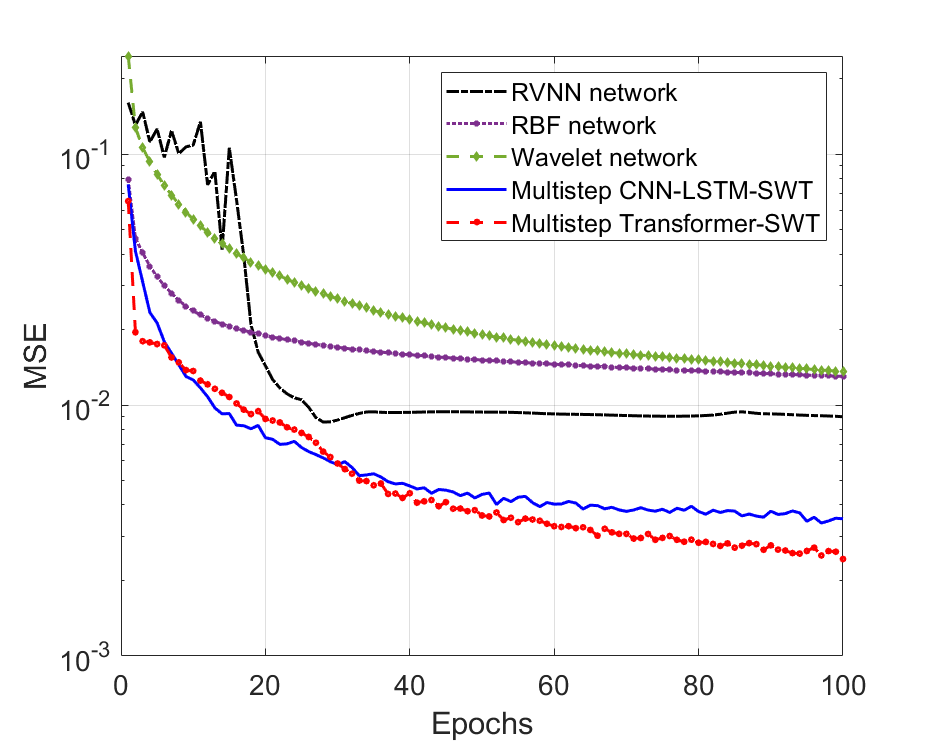}
\captionof{figure}{Learning curves of cascaded CNN-LSTM-SWT, cascaded transformer-SWT, MIMO-RVNN, wavelet network, and RBF network for 100 epochs}
\label{fig:curves}
\end{figure}

\begin{figure*}[t] 
\centering
\includegraphics[scale=.5]{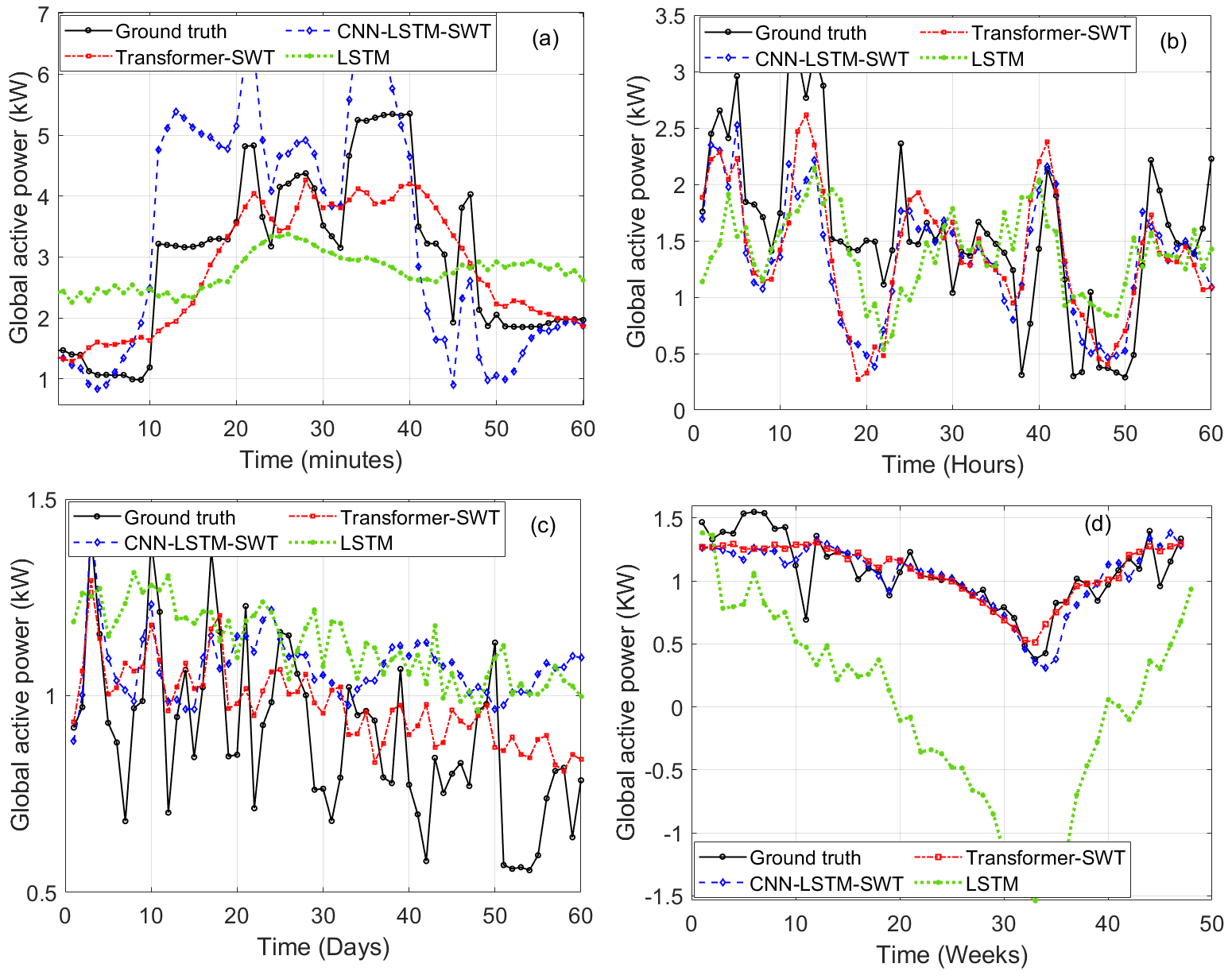}
\captionof{figure}{Performance comparison of the energy consumption prediction results for a) 60 minutes forecasting, b) 60 hours forecasting, c) 60 days forecasting, and d) 48 weeks forecasting}
\label{fig:a4}
\end{figure*}

Table \ref{table:2} shows the obtained results, along with other recently published models, LSTM and CNN-LSTM \citep{KIM201972}. 
The learning curves of the cascaded CNN-LSTM-SWT, cascaded transformer-SWT, MIMO-RVNN, wavelet network, and RBF network for 100 epochs are compared in Figure~\ref{fig:curves}. It can be observed that the proposed hybrid models achieve higher performance than previous methods. 
Figure~\ref{fig:a4} shows a selected timestamp from the validation set of the actual and forecasted 60-minutes of energy consumption using our deep learning-SWT hybrid models. The proposed models forecast energy consumption efficiently, by predicting the global and local features of energy consumption.

\begin{table}[t]
\centering
\footnotesize
\caption{Comparison results in terms of RMSE (kW)  of the proposed Multi-SWT-Deep Learning models with other existing models in the literature }\leavevmode\\
\begin{tabular}{@{}p{4.8cm}|p{1cm}p{1cm}p{1cm}p{1.2cm}@{}}
        \toprule
         Model & Minutely & Hourly & Daily  & Weekly \\
        \midrule
         CRBMs \citep{MOCANU201691}             & 0.9032   & 0.6906 & -      & 0.1822 \\
         LSTM-Seq2Seq   \citep{8945363}    & 0.6670   & 0.6250 & -      & -      \\
         FCRBM      \citep{MOCANU201691}        & 0.6663   & 0.6630 & -      & 0.1702 \\
         CNN-LSTM \citep{KIM201972}           & 0.6114   & 0.5957 & 0.3221 & 0.3085 \\
         CNN-BDLSTM    \citep{7793413}     & 0.5650   & -      & -      & -      \\
         CNN-GRU \citep{9141253}            & 0.4700   & -      & -      & -      \\
         CNN-LSTM-SWT   (proposed)  & 0.4645   & 0.5102 & 0.2793 & 0.1758 \\
         \textbf{Transformer-SWT (proposed)} & \textbf{0.3929} & \textbf{0.4183} & \textbf{0.2378} & \textbf{0.1574} \\
\bottomrule
\end{tabular}
\label{table:3}
\end{table}

\begin{table}[t]
\centering
\footnotesize
\caption{Evaluation of the most important feature in the daily case.}\leavevmode\\
\begin{tabular}{@{}p{3cm}|p{3.5cm}|p{2cm}p{2cm}p{2cm}@{}}
\toprule
Method      &Wavelet coefficients used              & RMSE (kW)   & MAE (kW)    & MAPE (\%) \\
\midrule
CNN-LSTM-SWT & All   &0.2793 	&0.2196	&31.73\\

~  & $A_3$   &0.2874 &	0.2270	&32.14\\
~  & $D_3$   &0.9459 &	0.9015	&100.13\\
~  & $D_2$   &0.9457 &	0.9015	&100.13\\
~  & $D_1$ &0.9458 &	0.9015	&100.14\\
Transformer-SWT & All    &0.2378 &	0.1832&	21.90\\
~ & $A_3$   &0.2501&	0.1903&	22.07\\
~ & $D_3$   &0.9878 &	0.9473&	99.69\\
~ & $D_2$   &0.9899&	0.9489&	99.77\\
~ & $D_1$  &0.9917&	0.9501&	99.87\\

\bottomrule
\end{tabular}
\end{table}

Figure~\ref{fig:a4}(a) demonstrates that the proposed approach can efficiently extract the irregular energy consumption patterns in the case of one-minute forecasting. Results for the hourly, daily and weekly cases are shown in Figure~\ref{fig:a4}(b),  Figure~\ref{fig:a4}(c) and  Figure~\ref{fig:a4}(d), respectively.  The results show that both models achieve superior performance to other models in household energy consumption. Table~\ref{table:3} shows the performance evaluation of the proposed technique in comparison with other recently proposed competitive methods. 

The experiment was set up with the same temporal resolution of minutely, hourly, daily, and weekly units. \cite{9141253} combined CNN with GRU to forecast time series energy consumption. \cite{KIM201972} proposed a hybrid model based on CNN and LSTM for the multivariate prediction of household energy consumption. \cite{MOCANU201691} proposed two stochastic models for time-series prediction of energy consumption, CRBM, and FCRBM. \cite{7793413} investigated a novel energy load prediction approach based on deep neural networks and an LSTM-based Sequence-to-Sequence (Seq2Seq). The results demonstrate that our transformer-based approach outperformed baseline power consumption prediction methods by a significant margin.

\begin{figure}[t]
\centering
\includegraphics[scale=.5]{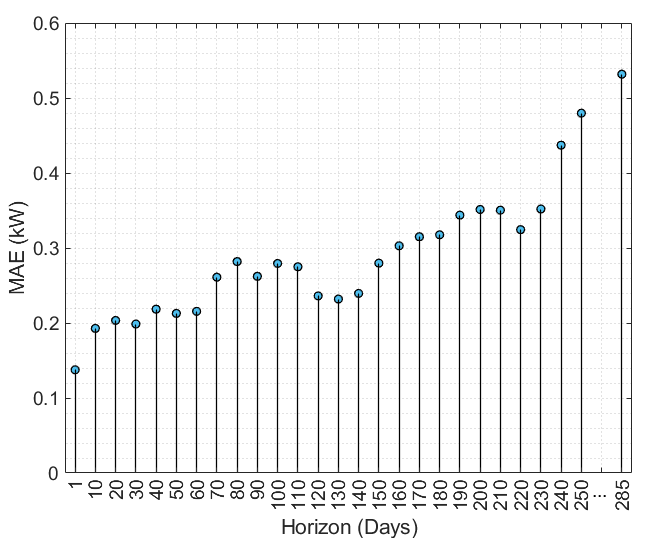}
\captionof{figure}{The evolution of MAE across the prediction horizon in daily case.}
\label{fig:mae}
\end{figure}

Next, we tested the performance of the proposed model across different prediction horizons, Figure~\ref{fig:mae} shows the MAE evolution across the whole testing data in the daily case. Taking the MAE of the next step forecasting as a reference, the model is 57\% less accurate at sample 60, 72\% less accurate at sample 120, 130\% less accurate at sample 180, 220\% less accurate at sample 240, and 287\% less at sample 285, the last sample in our testing dataset.  It should be noted that the average and the standard deviation of daily power consumption are $\mu=1.09kW$ and  $\sigma=0.41kW$.

Finally, we tested the significance of the different wavelet features. We generated daily predictions using individual wavelet features by setting the other wavelet coefficients to zero. Table 3 shows the obtained results. One can see that the most important feature is the third approximation subband $A_3$.

Despite the efficiency of the presented models, some limitations can be addressed in future works. First, the model was trained on a specific location, making  transfer learning only effective if applied to data with similar characteristics. We recommend training the model on additional data acquired at different sites and with different acquisition parameters to overcome this limitation. Second, the proposed models were trained and tested on noise-free and pre-processed data. Training and testing the proposed power consumption prediction methods on noisy and contaminated data can be considered in future work. 

\section{Conclusions}
In this study, we proposed and tested two hybrid methods based on the stationary wavelet transform (SWT) and deep learning for robust predictions of residential energy consumption. SWT is used to decompose electrical energy consumption time-series data and produce distinguishable feature representations needed for accurate predictions. Deep learning models such as convolutional-recurrent neural networks and deep transformers are used to process the SWT sub-band signals and characterize household energy consumption. 
Such accurate energy consumption forecasting methods pave the way for more reliable power supply, effective operation, and better sustainability of electricity generation systems. 

\section*{Acknowledgements}
This work has been funded by the ICT Fund, Telecommunications Regulatory Authority (TRA), PO.Box: 26662 Abu Dhabi, United Arab Emirates.

% \bibliographystyle{chicago}
% \bibliography{reference.bib}

\bibliographystyle{unsrtnat}
\bibliography{references}  %%% Uncomment this line and comment out the ``thebibliography'' section below to use the external .bib file (using bibtex) .

\begin{thebibliography}{46}
\providecommand{\natexlab}[1]{#1}
\providecommand{\url}[1]{\texttt{#1}}
\expandafter\ifx\csname urlstyle\endcsname\relax
  \providecommand{\doi}[1]{doi: #1}\else
  \providecommand{\doi}{doi: \begingroup \urlstyle{rm}\Url}\fi

\bibitem[Sajjad et~al.(2020)Sajjad, Khan, Ullah, Hussain, Ullah, Lee, and
  Baik]{9141253}
Muhammad Sajjad, Zulfiqar~Ahmad Khan, Amin Ullah, Tanveer Hussain, Waseem
  Ullah, Mi~Young Lee, and Sung~Wook Baik.
\newblock A novel {CNN-GRU-}based hybrid approach for short-term residential
  load forecasting.
\newblock \emph{IEEE Access}, 8:\penalty0 143759--143768, 2020.
\newblock \doi{10.1109/ACCESS.2020.3009537}.

\bibitem[{IEA}(2019)]{IEA}
{IEA}.
\newblock World energy outlook 2019, {IEA}, paris, 2019.
\newblock URL \url{https://www.iea.org/reports/world-energy-outlook-2019}.

\bibitem[Nejat et~al.(2015)Nejat, Jomehzadeh, Taheri, Gohari, and {Abd.
  Majid}]{NEJAT2015843}
Payam Nejat, Fatemeh Jomehzadeh, Mohammad~Mahdi Taheri, Mohammad Gohari, and
  Muhd~Zaimi {Abd. Majid}.
\newblock A global review of energy consumption, {CO} emissions and policy in
  the residential sector (with an overview of the top ten co2 emitting
  countries).
\newblock \emph{Renewable and Sustainable Energy Reviews}, 43:\penalty0
  843--862, 2015.
\newblock ISSN 1364-0321.
\newblock \doi{https://doi.org/10.1016/j.rser.2014.11.066}.
\newblock URL
  \url{https://www.sciencedirect.com/science/article/pii/S1364032114010053}.

\bibitem[Kim and Cho(2019)]{KIM201972}
Tae-Young Kim and Sung-Bae Cho.
\newblock Predicting residential energy consumption using {CNN-LSTM} neural
  networks.
\newblock \emph{Energy}, 182:\penalty0 72--81, 2019.
\newblock ISSN 0360-5442.
\newblock \doi{https://doi.org/10.1016/j.energy.2019.05.230}.
\newblock URL
  \url{https://www.sciencedirect.com/science/article/pii/S0360544219311223}.

\bibitem[Li et~al.(2017)Li, Ding, Zhao, Yi, and Zhang]{en10101525}
Chengdong Li, Zixiang Ding, Dongbin Zhao, Jianqiang Yi, and Guiqing Zhang.
\newblock Building energy consumption prediction: An extreme deep learning
  approach.
\newblock \emph{Energies}, 10\penalty0 (10):\penalty0 1--20, 2017.
\newblock ISSN 1996-1073.
\newblock \doi{10.3390/en10101525}.
\newblock URL \url{https://www.mdpi.com/1996-1073/10/10/1525}.

\bibitem[Siano(2014)]{SIANO2014461}
Pierluigi Siano.
\newblock Demand response and smart grids—a survey.
\newblock \emph{Renewable and Sustainable Energy Reviews}, 30:\penalty0
  461--478, 2014.
\newblock ISSN 1364-0321.
\newblock \doi{https://doi.org/10.1016/j.rser.2013.10.022}.
\newblock URL
  \url{https://www.sciencedirect.com/science/article/pii/S1364032113007211}.

\bibitem[Dong and Yang(2020)]{10.1007/978-981-13-9714-1_20}
Hongxiang Dong and Henan Yang.
\newblock {FPKC}: An efficient algorithm for improving short-term load
  forecasting.
\newblock In Jeng-Shyang Pan, Jianpo Li, Pei-Wei Tsai, and Lakhmi~C. Jain,
  editors, \emph{Advances in Intelligent Information Hiding and Multimedia
  Signal Processing}, pages 181--187, Singapore, 2020. Springer Singapore.
\newblock ISBN 978-981-13-9714-1.

\bibitem[Roldán-Blay et~al.(2013)Roldán-Blay, Escrivá-Escrivá, Álvarez
  Bel, Roldán-Porta, and Rodríguez-García]{ROLDANBLAY201338}
Carlos Roldán-Blay, Guillermo Escrivá-Escrivá, Carlos Álvarez Bel, Carlos
  Roldán-Porta, and Javier Rodríguez-García.
\newblock Upgrade of an artificial neural network prediction method for
  electrical consumption forecasting using an hourly temperature curve model.
\newblock \emph{Energy and Buildings}, 60:\penalty0 38--46, 2013.
\newblock ISSN 0378-7788.
\newblock \doi{https://doi.org/10.1016/j.enbuild.2012.12.009}.
\newblock URL
  \url{https://www.sciencedirect.com/science/article/pii/S0378778812006585}.

\bibitem[Ahmad et~al.(2014)Ahmad, Hassan, Abdullah, Rahman, Hussin, Abdullah,
  and Saidur]{AHMAD2014102}
A.S. Ahmad, M.Y. Hassan, M.P. Abdullah, H.A. Rahman, F.~Hussin, H.~Abdullah,
  and R.~Saidur.
\newblock A review on applications of {ANN} and {SVM} for building electrical
  energy consumption forecasting.
\newblock \emph{Renewable and Sustainable Energy Reviews}, 33:\penalty0
  102--109, 2014.
\newblock ISSN 1364-0321.
\newblock \doi{https://doi.org/10.1016/j.rser.2014.01.069}.
\newblock URL
  \url{https://www.sciencedirect.com/science/article/pii/S1364032114000914}.

\bibitem[Eskandari et~al.(2021)Eskandari, Imani, and
  Moghaddam]{ESKANDARI2021107173}
Hosein Eskandari, Maryam Imani, and Mohsen~Parsa Moghaddam.
\newblock Convolutional and recurrent neural network based model for short-term
  load forecasting.
\newblock \emph{Electric Power Systems Research}, 195:\penalty0 107173, 2021.
\newblock ISSN 0378-7796.
\newblock \doi{https://doi.org/10.1016/j.epsr.2021.107173}.
\newblock URL
  \url{https://www.sciencedirect.com/science/article/pii/S0378779621001541}.

\bibitem[xiang Zhao and Magoulès(2012)]{ZHAO20123586}
Hai xiang Zhao and Frédéric Magoulès.
\newblock A review on the prediction of building energy consumption.
\newblock \emph{Renewable and Sustainable Energy Reviews}, 16\penalty0
  (6):\penalty0 3586--3592, 2012.
\newblock ISSN 1364-0321.
\newblock \doi{https://doi.org/10.1016/j.rser.2012.02.049}.
\newblock URL
  \url{https://www.sciencedirect.com/science/article/pii/S1364032112001438}.

\bibitem[Kaboli et~al.(2016)Kaboli, Selvaraj, and Rahim]{KABOLI2016857}
S.~Hr.~Aghay Kaboli, J.~Selvaraj, and N.A. Rahim.
\newblock Long-term electric energy consumption forecasting via artificial
  cooperative search algorithm.
\newblock \emph{Energy}, 115:\penalty0 857--871, 2016.
\newblock ISSN 0360-5442.
\newblock \doi{https://doi.org/10.1016/j.energy.2016.09.015}.
\newblock URL
  \url{https://www.sciencedirect.com/science/article/pii/S0360544216312567}.

\bibitem[Kandananond(2011)]{en4081246}
Karin Kandananond.
\newblock Forecasting electricity demand in thailand with an artificial neural
  network approach.
\newblock \emph{Energies}, 4\penalty0 (8):\penalty0 1246--1257, 2011.
\newblock ISSN 1996-1073.
\newblock \doi{10.3390/en4081246}.
\newblock URL \url{https://www.mdpi.com/1996-1073/4/8/1246}.

\bibitem[Hadri et~al.(2021)Hadri, Najib, Bakhouya, Fakhri, and
  El~Arroussi]{en14185831}
Sarah Hadri, Mehdi Najib, Mohamed Bakhouya, Youssef Fakhri, and Mohamed
  El~Arroussi.
\newblock Performance evaluation of forecasting strategies for electricity
  consumption in buildings.
\newblock \emph{Energies}, 14\penalty0 (18), 2021.
\newblock ISSN 1996-1073.
\newblock \doi{10.3390/en14185831}.
\newblock URL \url{https://www.mdpi.com/1996-1073/14/18/5831}.

\bibitem[Jin et~al.(2022)Jin, Yang, Mo, Zeng, Zhou, Yan, and Ma]{JIN2022101442}
Ning Jin, Fan Yang, Yuchang Mo, Yongkang Zeng, Xiaokang Zhou, Ke~Yan, and Xiang
  Ma.
\newblock Highly accurate energy consumption forecasting model based on
  parallel {LSTM} neural networks.
\newblock \emph{Advanced Engineering Informatics}, 51:\penalty0 101442, 2022.
\newblock ISSN 1474-0346.
\newblock \doi{https://doi.org/10.1016/j.aei.2021.101442}.
\newblock URL
  \url{https://www.sciencedirect.com/science/article/pii/S1474034621001944}.

\bibitem[Somu et~al.(2020)Somu, {M R}, and Ramamritham]{SOMU2020114131}
Nivethitha Somu, Gauthama~Raman {M R}, and Krithi Ramamritham.
\newblock A hybrid model for building energy consumption forecasting using long
  short term memory networks.
\newblock \emph{Applied Energy}, 261:\penalty0 114131, 2020.
\newblock ISSN 0306-2619.
\newblock \doi{https://doi.org/10.1016/j.apenergy.2019.114131}.
\newblock URL
  \url{https://www.sciencedirect.com/science/article/pii/S0306261919318185}.

\bibitem[Khan et~al.(2021)Khan, Haq, Khan, Rho, Lee, and Baik]{KHAN2021107023}
Noman Khan, Ijaz~Ul Haq, Samee~Ullah Khan, Seungmin Rho, Mi~Young Lee, and
  Sung~Wook Baik.
\newblock {DB-N}et: A novel dilated {CNN} based multi-step forecasting model
  for power consumption in integrated local energy systems.
\newblock \emph{International Journal of Electrical Power \& Energy Systems},
  133:\penalty0 107023, 2021.
\newblock ISSN 0142-0615.
\newblock \doi{https://doi.org/10.1016/j.ijepes.2021.107023}.
\newblock URL
  \url{https://www.sciencedirect.com/science/article/pii/S0142061521002635}.

\bibitem[Hu et~al.(2020)Hu, Wang, and Lv]{HU2020598}
Huanling Hu, Lin Wang, and Sheng-Xiang Lv.
\newblock Forecasting energy consumption and wind power generation using deep
  echo state network.
\newblock \emph{Renewable Energy}, 154:\penalty0 598--613, 2020.
\newblock ISSN 0960-1481.
\newblock \doi{https://doi.org/10.1016/j.renene.2020.03.042}.
\newblock URL
  \url{https://www.sciencedirect.com/science/article/pii/S0960148120303645}.

\bibitem[Liu et~al.(2020)Liu, Tan, Xu, Chen, and Li]{LIU2020109675}
Tao Liu, Zehan Tan, Chengliang Xu, Huanxin Chen, and Zhengfei Li.
\newblock Study on deep reinforcement learning techniques for building energy
  consumption forecasting.
\newblock \emph{Energy and Buildings}, 208:\penalty0 109675, 2020.
\newblock ISSN 0378-7788.
\newblock \doi{https://doi.org/10.1016/j.enbuild.2019.109675}.
\newblock URL
  \url{https://www.sciencedirect.com/science/article/pii/S0378778819324740}.

\bibitem[Zhang et~al.(2020)Zhang, Tian, Li, Zhang, and Zuo]{ZHANG2020117531}
Guiqing Zhang, Chenlu Tian, Chengdong Li, Jun~Jason Zhang, and Wangda Zuo.
\newblock Accurate forecasting of building energy consumption via a novel
  ensembled deep learning method considering the cyclic feature.
\newblock \emph{Energy}, 201:\penalty0 117531, 2020.
\newblock ISSN 0360-5442.
\newblock \doi{https://doi.org/10.1016/j.energy.2020.117531}.
\newblock URL
  \url{https://www.sciencedirect.com/science/article/pii/S0360544220306381}.

\bibitem[Kazemzadeh et~al.(2020)Kazemzadeh, Amjadian, and
  Amraee]{KAZEMZADEH2020117948}
Mohammad-Rasool Kazemzadeh, Ali Amjadian, and Turaj Amraee.
\newblock A hybrid data mining driven algorithm for long term electric peak
  load and energy demand forecasting.
\newblock \emph{Energy}, 204:\penalty0 117948, 2020.
\newblock ISSN 0360-5442.
\newblock \doi{https://doi.org/10.1016/j.energy.2020.117948}.
\newblock URL
  \url{https://www.sciencedirect.com/science/article/pii/S0360544220310550}.

\bibitem[Lu et~al.(2020)Lu, Cheng, Ma, and Hu]{LU2020117756}
Hongfang Lu, Feifei Cheng, Xin Ma, and Gang Hu.
\newblock Short-term prediction of building energy consumption employing an
  improved extreme gradient boosting model: A case study of an intake tower.
\newblock \emph{Energy}, 203:\penalty0 117756, 2020.
\newblock ISSN 0360-5442.
\newblock \doi{https://doi.org/10.1016/j.energy.2020.117756}.
\newblock URL
  \url{https://www.sciencedirect.com/science/article/pii/S036054422030863X}.

\bibitem[Yan et~al.(2019)Yan, Li, Ji, Qi, and Du]{8880605}
Ke~Yan, Wei Li, Zhiwei Ji, Meng Qi, and Yang Du.
\newblock A hybrid {LSTM} neural network for energy consumption forecasting of
  individual households.
\newblock \emph{IEEE Access}, 7:\penalty0 157633--157642, 2019.
\newblock \doi{10.1109/ACCESS.2019.2949065}.

\bibitem[Vaswani et~al.(2017)Vaswani, Shazeer, Parmar, Uszkoreit, Jones, Gomez,
  Kaiser, and Polosukhin]{NIPS2017_3f5ee243}
Ashish Vaswani, Noam Shazeer, Niki Parmar, Jakob Uszkoreit, Llion Jones,
  Aidan~N Gomez, \L~ukasz Kaiser, and Illia Polosukhin.
\newblock Attention is all you need.
\newblock In I.~Guyon, U.~Von Luxburg, S.~Bengio, H.~Wallach, R.~Fergus,
  S.~Vishwanathan, and R.~Garnett, editors, \emph{Advances in Neural
  Information Processing Systems}, volume~30. Curran Associates, Inc., 2017.
\newblock URL
  \url{https://proceedings.neurips.cc/paper/2017/file/3f5ee243547dee91fbd053c1c4a845aa-Paper.pdf}.

\bibitem[Wu et~al.(2020)Wu, Green, Ben, and
  O'Banion]{https://doi.org/10.48550/arxiv.2001.08317}
Neo Wu, Bradley Green, Xue Ben, and Shawn O'Banion.
\newblock Deep transformer models for time series forecasting: The influenza
  prevalence case, 2020.
\newblock URL \url{https://arxiv.org/abs/2001.08317}.

\bibitem[Li et~al.(2019)Li, Jin, Xuan, Zhou, Chen, Wang, and
  Yan]{NEURIPS2019_6775a063}
Shiyang Li, Xiaoyong Jin, Yao Xuan, Xiyou Zhou, Wenhu Chen, Yu-Xiang Wang, and
  Xifeng Yan.
\newblock Enhancing the locality and breaking the memory bottleneck of
  transformer on time series forecasting.
\newblock In H.~Wallach, H.~Larochelle, A.~Beygelzimer, F.~d\textquotesingle
  Alch\'{e}-Buc, E.~Fox, and R.~Garnett, editors, \emph{Advances in Neural
  Information Processing Systems}, volume~32. Curran Associates, Inc., 2019.
\newblock URL
  \url{https://proceedings.neurips.cc/paper/2019/file/6775a0635c302542da2c32aa19d86be0-Paper.pdf}.

\bibitem[Lin et~al.(2020)Lin, Koprinska, and
  Rana]{10.1007/978-3-030-63836-8_51}
Yang Lin, Irena Koprinska, and Mashud Rana.
\newblock Springnet: Transformer and spring {DTW} for time series forecasting.
\newblock In Haiqin Yang, Kitsuchart Pasupa, Andrew Chi-Sing Leung, James~T.
  Kwok, Jonathan~H. Chan, and Irwin King, editors, \emph{Neural Information
  Processing}, pages 616--628, Cham, 2020. Springer International Publishing.
\newblock ISBN 978-3-030-63836-8.

\bibitem[Saad~Saoud et~al.(2022)Saad~Saoud, Al-Marzouqi, and Hussein]{9672113}
Lyes Saad~Saoud, Hasan Al-Marzouqi, and Ramy Hussein.
\newblock Household energy consumption prediction using the stationary wavelet
  transform and transformers.
\newblock \emph{IEEE Access}, 10:\penalty0 5171--5183, 2022.
\newblock \doi{10.1109/ACCESS.2022.3140818}.

\bibitem[Nason and Silverman(1995)]{Nason1995}
G.~P. Nason and B.~W. Silverman.
\newblock \emph{The Stationary Wavelet Transform and some Statistical
  Applications}, pages 281--299.
\newblock Springer New York, New York, NY, 1995.
\newblock ISBN 978-1-4612-2544-7.
\newblock \doi{10.1007/978-1-4612-2544-7_17}.
\newblock URL \url{https://doi.org/10.1007/978-1-4612-2544-7_17}.

\bibitem[Supratid et~al.(2017)Supratid, Aribarg, and Supratid]{article}
Siriporn Supratid, Thannob Aribarg, and Seree Supratid.
\newblock An integration of stationary wavelet transform and nonlinear
  autoregressive neural network with exogenous input for baseline and future
  forecasting of reservoir inflow.
\newblock \emph{Water Resources Management}, 31:\penalty0 1--21, 09 2017.
\newblock \doi{10.1007/s11269-017-1726-2}.

\bibitem[İlker Güven and Şimşir(2020)]{GUVEN2020106678}
İlker Güven and Fuat Şimşir.
\newblock Demand forecasting with color parameter in retail apparel industry
  using artificial neural networks {(ANN)} and support vector machines {(SVM)}
  methods.
\newblock \emph{Computers \& Industrial Engineering}, 147:\penalty0 106678,
  2020.
\newblock ISSN 0360-8352.
\newblock \doi{https://doi.org/10.1016/j.cie.2020.106678}.
\newblock URL
  \url{https://www.sciencedirect.com/science/article/pii/S0360835220304125}.

\bibitem[Hadiyan et~al.(2020)Hadiyan, Moeini, and
  Ehsanzadeh]{HADIYAN2020100401}
Pedram~Pishgah Hadiyan, Ramtin Moeini, and Eghbal Ehsanzadeh.
\newblock Application of static and dynamic artificial neural networks for
  forecasting inflow discharges, case study: Sefidroud dam reservoir.
\newblock \emph{Sustainable Computing: Informatics and Systems}, 27:\penalty0
  100401, 2020.
\newblock ISSN 2210-5379.
\newblock \doi{https://doi.org/10.1016/j.suscom.2020.100401}.
\newblock URL
  \url{https://www.sciencedirect.com/science/article/pii/S2210537920301281}.

\bibitem[Zhou et~al.(2020)Zhou, Liu, and Duan]{ZHOU2020125127}
Fanhan Zhou, Bingjun Liu, and Kai Duan.
\newblock Coupling wavelet transform and artificial neural network for
  forecasting estuarine salinity.
\newblock \emph{Journal of Hydrology}, 588:\penalty0 125127, 2020.
\newblock ISSN 0022-1694.
\newblock \doi{https://doi.org/10.1016/j.jhydrol.2020.125127}.
\newblock URL
  \url{https://www.sciencedirect.com/science/article/pii/S0022169420305874}.

\bibitem[Talaat et~al.(2020)Talaat, Farahat, Mansour, and
  Hatata]{TALAAT2020117087}
M.~Talaat, M.A. Farahat, Noura Mansour, and A.Y. Hatata.
\newblock Load forecasting based on grasshopper optimization and a multilayer
  feed-forward neural network using regressive approach.
\newblock \emph{Energy}, 196:\penalty0 117087, 2020.
\newblock ISSN 0360-5442.
\newblock \doi{https://doi.org/10.1016/j.energy.2020.117087}.
\newblock URL
  \url{https://www.sciencedirect.com/science/article/pii/S0360544220301948}.

\bibitem[Abadi et~al.(2016)Abadi, Barham, Chen, Chen, Davis, Dean, Devin,
  Ghemawat, Irving, Isard, Kudlur, Levenberg, Monga, Moore, Murray, Steiner,
  Tucker, Vasudevan, Warden, Wicke, Yu, and Zheng]{199317}
Mart{\'\i}n Abadi, Paul Barham, Jianmin Chen, Zhifeng Chen, Andy Davis, Jeffrey
  Dean, Matthieu Devin, Sanjay Ghemawat, Geoffrey Irving, Michael Isard,
  Manjunath Kudlur, Josh Levenberg, Rajat Monga, Sherry Moore, Derek~G. Murray,
  Benoit Steiner, Paul Tucker, Vijay Vasudevan, Pete Warden, Martin Wicke, Yuan
  Yu, and Xiaoqiang Zheng.
\newblock {TensorFlow}: A system for {Large-Scale} machine learning.
\newblock In \emph{12th USENIX Symposium on Operating Systems Design and
  Implementation (OSDI 16)}, pages 265--283, Savannah, GA, November 2016.
  USENIX Association.
\newblock ISBN 978-1-931971-33-1.
\newblock URL
  \url{https://www.usenix.org/conference/osdi16/technical-sessions/presentation/abadi}.

\bibitem[Hochreiter and Schmidhuber(1997)]{10.1162/neco.1997.9.8.1735}
Sepp Hochreiter and Jürgen Schmidhuber.
\newblock {Long Short-Term Memory}.
\newblock \emph{Neural Computation}, 9\penalty0 (8):\penalty0 1735--1780, 11
  1997.
\newblock ISSN 0899-7667.
\newblock \doi{10.1162/neco.1997.9.8.1735}.
\newblock URL \url{https://doi.org/10.1162/neco.1997.9.8.1735}.

\bibitem[Kazemi et~al.(2019)Kazemi, Goel, Eghbali, Ramanan, Sahota, Thakur, Wu,
  Smyth, Poupart, and Brubaker]{https://doi.org/10.48550/arxiv.1907.05321}
Seyed~Mehran Kazemi, Rishab Goel, Sepehr Eghbali, Janahan Ramanan, Jaspreet
  Sahota, Sanjay Thakur, Stella Wu, Cathal Smyth, Pascal Poupart, and Marcus
  Brubaker.
\newblock {Time2Vec}: Learning a vector representation of time, 2019.
\newblock URL \url{https://arxiv.org/abs/1907.05321}.

\bibitem[Shen et~al.(2020)Shen, Jiang, Wang, Jin, and Cheng]{9194535}
Yinghan Shen, Xuhui Jiang, Yuanzhuo Wang, Xiaolong Jin, and Xueqi Cheng.
\newblock Dynamic relation extraction with a learnable temporal encoding
  method.
\newblock In \emph{2020 IEEE International Conference on Knowledge Graph
  (ICKG)}, pages 235--242, 2020.
\newblock \doi{10.1109/ICBK50248.2020.00042}.

\bibitem[Hebrail(2012)]{misc_individual_household_electric_power_consumption_235}
Alice Hebrail, Georges \&~Berard.
\newblock {Individual household electric power consumption}.
\newblock UCI Machine Learning Repository, 2012.

\bibitem[Mocanu et~al.(2016)Mocanu, Nguyen, Gibescu, and Kling]{MOCANU201691}
Elena Mocanu, Phuong~H. Nguyen, Madeleine Gibescu, and Wil~L. Kling.
\newblock Deep learning for estimating building energy consumption.
\newblock \emph{Sustainable Energy, Grids and Networks}, 6:\penalty0 91--99,
  2016.
\newblock ISSN 2352-4677.
\newblock \doi{https://doi.org/10.1016/j.segan.2016.02.005}.
\newblock URL
  \url{https://www.sciencedirect.com/science/article/pii/S2352467716000163}.

\bibitem[Marino et~al.(2016)Marino, Amarasinghe, and Manic]{7793413}
Daniel~L. Marino, Kasun Amarasinghe, and Milos Manic.
\newblock Building energy load forecasting using deep neural networks.
\newblock In \emph{IECON 2016 - 42nd Annual Conference of the IEEE Industrial
  Electronics Society}, pages 7046--7051, 2016.
\newblock \doi{10.1109/IECON.2016.7793413}.

\bibitem[Ullah et~al.(2020)Ullah, Ullah, Haq, Rho, and Baik]{8945363}
Fath U~Min Ullah, Amin Ullah, Ijaz~Ul Haq, Seungmin Rho, and Sung~Wook Baik.
\newblock Short-term prediction of residential power energy consumption via
  {CNN} and multi-layer bi-directional {LSTM} networks.
\newblock \emph{IEEE Access}, 8:\penalty0 123369--123380, 2020.
\newblock \doi{10.1109/ACCESS.2019.2963045}.

\bibitem[Brownlee(2017)]{brownlee2017introduction}
J.~Brownlee.
\newblock \emph{Introduction to Time Series Forecasting With Python: How to
  Prepare Data and Develop Models to Predict the Future}.
\newblock Machine Learning Mastery, 2017.
\newblock URL \url{https://books.google.ae/books?id=-AiqDwAAQBAJ}.

\bibitem[Ahmed et~al.(2010)Ahmed, Atiya, Gayar, and
  El-Shishiny]{doi:10.1080/07474938.2010.481556}
Nesreen~K. Ahmed, Amir~F. Atiya, Neamat~El Gayar, and Hisham El-Shishiny.
\newblock An empirical comparison of machine learning models for time series
  forecasting.
\newblock \emph{Econometric Reviews}, 29\penalty0 (5-6):\penalty0 594--621,
  2010.
\newblock \doi{10.1080/07474938.2010.481556}.
\newblock URL \url{https://doi.org/10.1080/07474938.2010.481556}.

\bibitem[Shin and Woo(2022)]{en15134880}
Sun-Youn Shin and Han-Gyun Woo.
\newblock Energy consumption forecasting in korea using machine learning
  algorithms.
\newblock \emph{Energies}, 15\penalty0 (13), 2022.
\newblock ISSN 1996-1073.
\newblock \doi{10.3390/en15134880}.
\newblock URL \url{https://www.mdpi.com/1996-1073/15/13/4880}.

\bibitem[Nazir et~al.(2021)Nazir, Aziz, Hosen, Aziz, and Murthy]{Nazir_2021}
S.~Nazir, Azlan~Ab Aziz, J.~Hosen, Nor~Azlina Aziz, and G.~Ramana Murthy.
\newblock Forecast energy consumption time-series dataset using multistep
  {LSTM} models.
\newblock \emph{Journal of Physics: Conference Series}, 1933\penalty0
  (1):\penalty0 012054, jun 2021.
\newblock \doi{10.1088/1742-6596/1933/1/012054}.
\newblock URL \url{https://doi.org/10.1088/1742-6596/1933/1/012054}.

\end{thebibliography}

\end{document}